\documentclass[journal]{IEEEtran}

\usepackage{xcolor}
\newcommand{\revised}[1]{\textcolor{black}{#1}}

\usepackage{amsmath,amsfonts}

\usepackage{bm}
\usepackage{algorithmic}
\usepackage{algorithm}

\usepackage[caption=false]{subfig}
\usepackage{array}
\usepackage{textcomp}
\usepackage{stfloats}
\usepackage{url}
\usepackage{verbatim}
\usepackage[pdftex]{graphicx}
\graphicspath{./figs/}

\usepackage{booktabs}
\usepackage{tabularx}

\usepackage{cite}

\makeatletter
\let\NAT@parse\undefined
\makeatother

\usepackage[colorlinks,linkcolor=black,anchorcolor=black,citecolor=blue,urlcolor=blue,hyperfootnotes=true]{hyperref}   
\usepackage[all]{hypcap} 
\begin{document}
\bstctlcite{IEEEtran:BSTcontrol}

%
\title{Servo Integrated Nonlinear Model Predictive Control for Overactuated Tiltable-Quadrotors}

%

\author{Jinjie~Li$^{1}$,~\IEEEmembership{Graduate Student Member,~IEEE~RAS}, Junichiro~Sugihara$^{2}$, and~Moju~Zhao$^{1}$,~\IEEEmembership{Member,~IEEE}
\thanks{Manuscript received: April 5, 2024; Revised: July 6, 2024; Accepted: August 9, 2024. This letter was recommended for publication by Editor Jaydev P. Desai upon evaluation of the Reviewers’ comments. \textit{(Corresponding author: Moju Zhao)}}
\thanks{$^{1}$J. Li, and M. Zhao are with the DRAGON Lab at the Department of Mechanical Engineering, the University of Tokyo, Tokyo, 113-8654, Japan {\tt\small \{jinjie-li, chou\}@dragon.t.u-tokyo.ac.jp}}
\thanks{$^{2}$J. Sugihara is with the JSK Lab at the Department of Mechano-Informatics, the University of Tokyo, Tokyo, 113-8654, Japan {\tt\small j-sugihara@jsk.imi.i.u-tokyo.ac.jp}}
\thanks{Digital Object Identifier (DOI): see top of this page.}
\vspace{-0.5cm}
}

%
%


\markboth{IEEE Robotics and Automation Letters. Preprint Version. Accepted August, 2024}%
{Li \MakeLowercase{\textit{et al.}}: Servo Integrated NMPC for Overactuated Tiltable-Quadrotors}

%



\maketitle

\begin{abstract}
\revised{Utilizing a servo to tilt each rotor transforms quadrotors from underactuated to overactuated systems, allowing for independent control of both attitude and position, which provides advantages for aerial manipulation. However, this enhancement also introduces model nonlinearity, sluggish servo response, and limited operational range into the system, posing challenges to dynamic control.}
In this study, we propose a control approach for tiltable-quadrotors based on nonlinear model predictive control (NMPC). Unlike conventional cascade methods, our approach preserves the full dynamics without simplification. It directly uses rotor thrust and servo angle as control inputs, \revised{where their limited working ranges are considered input constraints.} Notably, we incorporate a first-order servo model within the NMPC framework. 
\revised{Simulation reveals that integrating the servo dynamics is not only an enhancement to control performance but also a critical factor for optimization convergence.}
To evaluate the effectiveness of our approach, we fabricate a tiltable-quadrotor and deploy the algorithm onboard at 100 Hz. Extensive real-world experiments demonstrate rapid, robust, and smooth pose-tracking performance.
\end{abstract}


%
\IEEEpeerreviewmaketitle


\section{Introduction}
%
%
%
%

\IEEEPARstart{A}{erial} robots have increasingly attracted attention
due to their mobility in three-dimensional space.
As the most popular aerial robot, quadrotors have been applied in various areas including transportation, inspection, and search \& rescue \cite{floreano_science_2015}.
Traditional quadrotors are \textit{underactuated}, meaning that the number of independent control inputs (i.e., the rotation speed of four vertically oriented rotors) is fewer than the required six degrees of freedom (DoF) for full-pose motion, leading to \revised{the independent control of only four flat outputs: three-axis position and yaw angle.}
However, numerous applications require independent attitude control, particularly in the field of aerial manipulation \cite{ollero_past_2022}.



\setlength{\textfloatsep}{8pt plus 1.0pt minus 2.0pt}
\begin{figure}[t]
    \centerline{\includegraphics[trim=0 0 0 0,clip,width=3.5in]{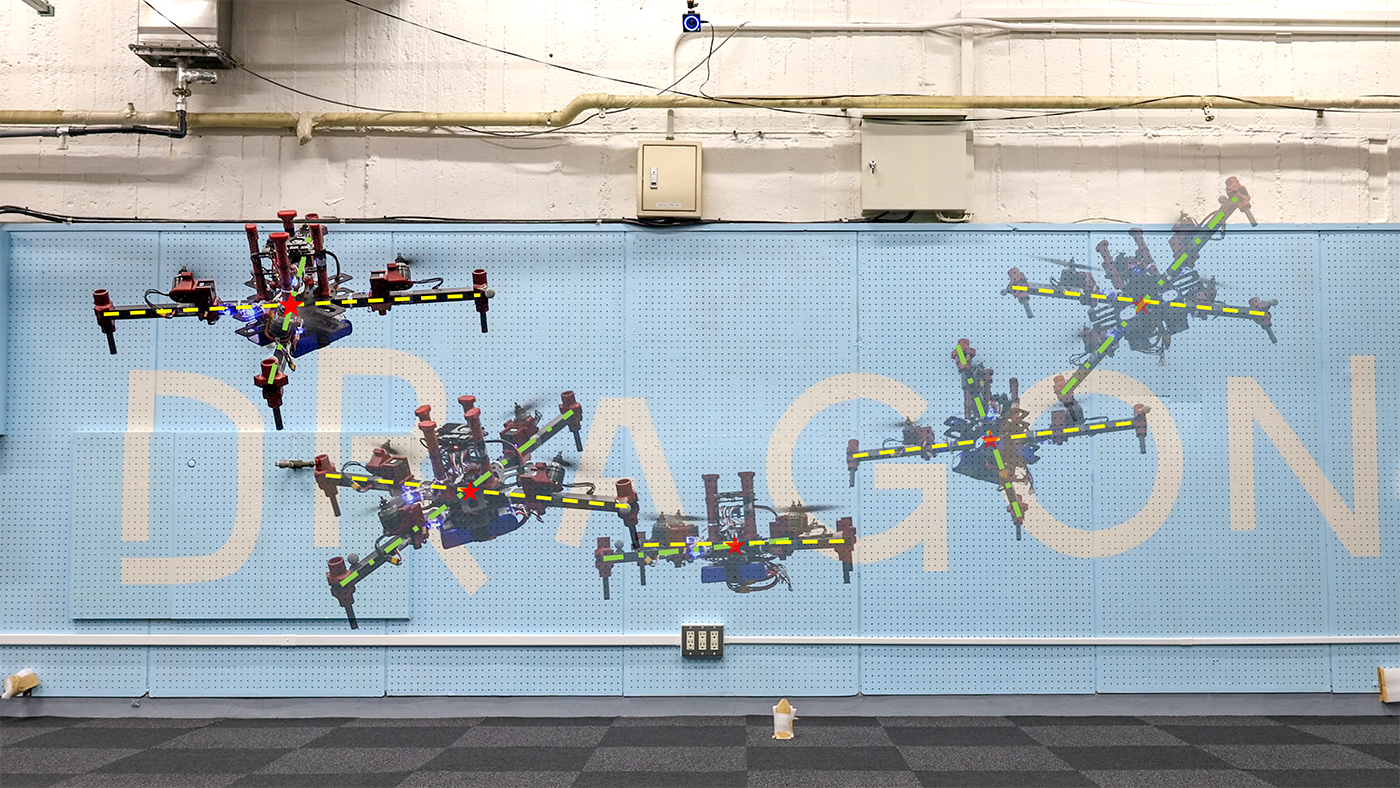}} 
    \vspace*{-1mm}
    \caption{Our self-made tiltable-quadrotor is tracking a pose trajectory based on the proposed NMPC method. It demonstrates the capability for independent control in position and attitude to track different Lemniscate curves.}
    \label{fig:flight}
\end{figure}

To overcome the challenges associated with \textit{underactuation}, researchers have developed two primary ways to achieve \textit{fullactuation} or \textit{overactuation} in rotor-based aerial robots.
One way involves adding rotors with fixed tilting angles \cite{brescianini_design_2016, flores_robust_2022}, named as \textit{fixed-rotor} robots. While these robots maintain mechanical and control simplicity comparable to their underactuated counterparts, they encounter issues like constant internal forces, reduced efficiency, and constrained wrench generation in certain directions.
Alternatively, the \textit{tilt-rotor} design introduces servo modules to vary thrust directions, \revised{reducing internal force} and enhancing energy efficiency. 
\revised{If we aim to enable conventional quadrotors to control independently both position and attitude while maintaining the symmetry of roll and pitch, a relatively straightforward idea is to add a servo-tilting mechanism to each rotor, allowing it to rotate along the arm.}
Therefore, our research focuses on the tiltable-quadrotor, as exemplified in Fig. \ref{fig:flight}.

The first real-world flight of a quadrotor equipped with tiltable rotors was achieved by Ryll et al., where the control approach is detailed in \cite{ryll_novel_2015}. Subsequently, the Voliro project \cite{kamel_voliro_2018} introduced a hexacopter with tiltable rotors, marking a pioneering step towards real-world omnidirectional flight.
In addition, Senkul et al. \cite{senkul_system_2016} developed a quadrotor capable of tilting its rotors along two axes, but they assumed the same thrust generation across all rotors, restricting the vehicle's maneuverability.
Despite these advantages, flight control for tiltable-multirotors remains challenging.

The control complexity of tiltable-multirotors arises primarily from two factors.
First, the introduction of additional DoFs to vary tilting angles significantly increases the system's nonlinearity.
These extra DoFs make the system overactuated, resulting in multiple solutions in control allocation.
Second, the inherent dynamics of tilting servos impede prompt motion control for agile trajectory tracking.
This slow property can be caused by dead time or the servo's time constant,
which has been reported as a critical factor in flight stability \cite{ryll_novel_2015}. Similar challenges have been observed in other servo-equipped aerial robots, such as SPIDAR \cite{zhao_design_2023} and Perching Arm \cite{nishio_design_2024}.

\begin{figure}[bt] 
    \centering
    \subfloat[Workflow in \cite{ryll_novel_2015, scholz_model_2016, kamel_voliro_2018, allenspach_design_2020, brunner_trajectory_2020}]{
        \includegraphics[trim=6.3cm 10.5cm 6.5cm 6cm,clip,width=3.4in]{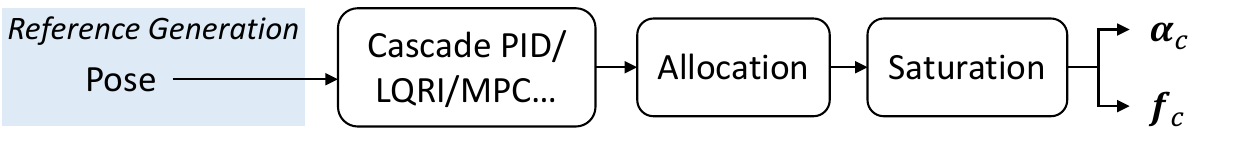}
        \label{fig:workflow_pre}
    } \\ \vspace{-3mm}
    \subfloat[Our workflow.]{
        \includegraphics[trim=6.3cm 6cm 6.5cm 9.5cm,clip,width=3.4in]{figs/workflow_comparison_v3.pdf}
        \label{fig:workflow_ours}
    }
    \caption{Comparative analysis of the previous and proposed workflows. Unlike the previous cascade structure, the proposed method directly integrates constraints and allocation within the NMPC optimization.
    \revised{Furthermore, servos are explicitly modeled. We directly use the servo angle and thrust as commands, requiring no extra integrator as in other research \cite{bicego_nonlinear_2020}.}}
    \label{fig:workflow_comparison}
\end{figure}

To address the first challenge, plenty of research has simplified this complex control problem by decoupling it into control and allocation components, as illustrated in Fig. \ref{fig:workflow_pre}. For control, various strategies have been proposed based on established control theories, including feedback linearization \cite{ryll_novel_2015}, nonlinear inverse dynamics \cite{scholz_model_2016}, cascade PID \cite{kamel_voliro_2018}, LQRI \cite{allenspach_design_2020}, and NMPC \cite{brunner_trajectory_2020}. 
For allocation, Moore–Penrose inverse is typically utilized to map control inputs \revised{(for most cases, wrench)} to specific actuator commands.
Although these approaches demonstrate effectiveness in real-world flights, their cascade structure may not fully leverage the potential of overactuated systems or elegantly address actuators' delay and constraints.
To overcome these drawbacks, recent studies have started to integrate control and allocation within a single optimization framework, notably through nonlinear model predictive control (NMPC).
This integrated approach has shown promise in conventional quadrotors \cite{sun_comparative_2022, li_nonlinear_2023} and is gaining traction for tiltable-multirotors. For instance, Bicego et al. \cite{bicego_nonlinear_2020} proposed an NMPC framework suitable for various multi-rotor designs, tested on a hexrotor that all motors can tilt for the same angle. However, its applicability to drones with independently tiltable rotors remains unverified. Shawky et al. \cite{shawky_nonlinear_2021} developed an NMPC controller for tiltable hexacopters, albeit only validated in Gazebo simulation.
Despite these advancements, the deployment of unified NMPC in real-world tiltable-multirotors is scarcely reported, and thus one highlight in this work is the implementation on a real tiltable-quadrotor.


To address the second challenge, some studies \cite{kamel_voliro_2018, shawky_nonlinear_2021} have chosen to overlook the servo effect, which consequently reduces the control performance. Ryll et al. \cite{ryll_novel_2015} implemented a Smith predictor to address the slow servo response. Although effective, this method is complicated since conventional feedback controllers lack the predictive property.
\revised{In contrast, NMPC inherently incorporates prediction, offering an advantageous framework for systems with sluggish actuators. For example, the works \cite{yigit_dynamic_2023}, \cite{cuvillon_offset-free_2023} leverage NMPC to control an aerial manipulator comprising both an agile multirotor and relatively slow carriers/winches.}
For tiltable multirotors, one approach to indirectly consider servomechanism within NMPC is to constrain the change rate of the resultant wrench, a method proposed by Brunner et al. \cite{brunner_trajectory_2020}. While this technique helps manage the servo dynamics, it cannot accurately capture the entire scope of servo behavior. More recent studies \cite{bicego_nonlinear_2020, brunner_mpc_2022} integrate the actuator model more explicitly by using their derivatives as control inputs \revised{(so-called ``delta-input formulation" \cite{yigit_dynamic_2023})} and trying to constrain the range of these derivatives.
\revised{Although the ``$\Delta u$ formulation" can describe various systems including first-order models, this generality might be unnecessary if the system is already known to be first-order. Furthermore, this way requires an additional integrator to accurately obtain the control command, and the actuators' measurement noise presents challenges for identifying their derivatives' boundaries in practice. In contrast, we model the servo as a first-order system, eliminating the need for an integrator and simplifying the identification process.}

In this work, we introduce a unified nonlinear model predictive control approach for tiltable-quadrotors, as illustrated in \ref{fig:workflow_ours}.
This approach leverages allocation as an external reference, rather than a module within the control loop.
Our method fully incorporates the nonlinear dynamics into the NMPC framework, directly utilizing rotor thrust and servo angle as control inputs.
Furthermore, we explicitly integrate the servo \revised{as a first-order system} within the model. We find that the inclusion of servo dynamics not only improves model fidelity but also facilitates optimization convergence.

\setlength{\dbltextfloatsep}{8pt plus 1.0pt minus 2.0pt}

The main contributions of this article are:
\begin{enumerate}
\item We propose a \revised{servo-integrated} NMPC framework for tiltable-quadrotors, which \revised{explicitly considers the servo as a first-order system. \revised{This framework has no simplification of the nonlinear robot model, resulting in the full exploration of the hardware potential.}}

\item \revised{We point out that due to the servo angle's nonlinearity and working range, considering the servo model is more critical than thrust for flight performance. In addition, modeling the servo is crucial for optimization convergence. These findings are verified in simulations.}

\item \revised{We thoroughly evaluate the real-world control performance on various  references with a self-developed tiltable-quadrotor, 
where the NMPC controller is running at 100 Hz on an onboard computer.} To the best of the authors' knowledge, this is the first time an actuator-level NMPC is implemented on a real tiltable-quadrotor.
\end{enumerate}

The remainder of the article is organized as follows. The modeling for the tiltable-quadrotor is introduced in Sec. \ref{sec:modeling}. The control approach is presented in Sec. \ref{sec:control}, followed by simulation analysis in Sec. \ref{sec:simulation}. We then show the experimental results in Sec. \ref{sec:experiments} and finally the conclusion in Sec. \ref{sec:conclusion}.

\section{Modeling} \label{sec:modeling}

\subsection{Coordinate Systems and Notation}

\setlength{\textfloatsep}{8pt plus 1.0pt minus 2.0pt}
\begin{figure}[b]
    \centerline{\includegraphics[trim=7.5cm 5cm 7.5cm 4.5cm,clip,width=3.4in]{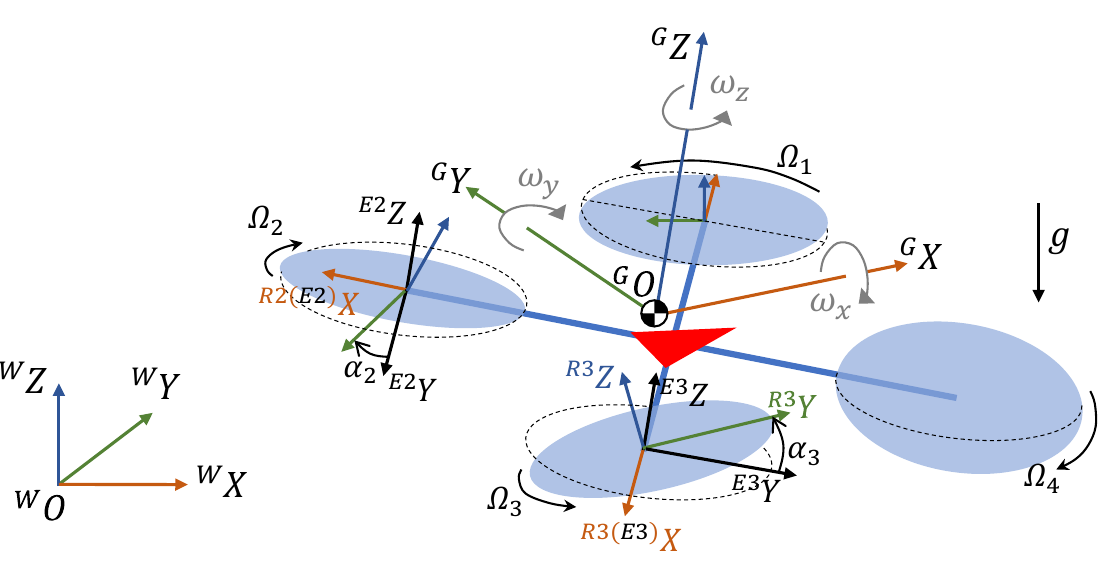}} 
    \vspace*{-3mm}
    \caption{Diagram of a tiltable-quadrotor with the ENU (X East, Y North, Z Up) inertial frame and the FLU (X Forward, Y Left, Z Up) body frame.}
    \label{fig:coordinate}
\end{figure}

We denote scalars in unbold $x, X \in \mathbb{R}$, vectors in bold lowercase $\boldsymbol{x} \in \mathbb{R}^n$, and matrices in bold uppercase $\boldsymbol{X} \in \mathbb{R}^{n \times m}$.
We use $\hat{\cdot}$ to denote estimated values. A vector in $\left\{\mathcal{W}\right\}$ can be denoted as $^{W}\boldsymbol{p}$, and the rotation from $\left\{\mathcal{G}\right\}$ to $\left\{\mathcal{W}\right\}$ is denoted as $^{W}_{G}\boldsymbol{R}$ (rotation matrix)  or $^{W}_{G}\boldsymbol{q}=\left[q_w, q_x, q_y, q_z\right]^T$ (attitude quaternion).

Depicted in Fig. \ref{fig:coordinate}, the coordinate systems contain the world inertial frame $\left\{\mathcal{W}\right\}$, the body frame $\left\{\mathcal{G}\right\}$ whose origin is at the center of gravity (CoG), as well as the $i$th arm-end frame $\left\{\mathcal{E}_i\right\}$ and rotor frame $\left\{\mathcal{R}_i\right\}$ $(i=1,2,3,4)$. The origin of $\left\{\mathcal{E}_i\right\}$ is positioned at the extremity of the $i$th arm, where its X axis points outward, and its Z axis aligns parallel to the Z axis of $\left\{\mathcal{G}\right\}$. Frame $\left\{\mathcal{R}_i\right\}$ is derived from $\left\{\mathcal{E}_i\right\}$ through a rotation about the X axis, with the rotation angle denoted as $\alpha_i$.


\subsection{Tiltable-Quadrotor Model}

The model for a tiltable-quadrotor has several parts. First, we establish the rotor model and get its resultant wrench. Then we introduce a first-order model to explicitly describe the servo dynamics. Finally, we apply the wrench to the rigid-body dynamics to obtain the full robot model.

\subsubsection{Rotor Model}

Motors rotate their propellers to generate thrust and torque. Here we neglect the angular acceleration of rotors and adopt commonly used quadratic fit to describe the motor dynamics
\begin{equation}
\label{eq:rotor_dynamics}
    f_i={k}_{t} \ \Omega_i^2, \quad \tau_i={k}_{q} \ \Omega_i^2,
\end{equation}
where $\Omega_i$ is the rotating speed of the $i$th propeller, ${k}_{t}$ and ${k}_{q}$ are coefficients identified from experiments. It is assumed that the motors are able to achieve the rotation speed $\Omega_i$ with negligible transients.




\subsubsection{\revised{Resultant Wrench}}

\revised{To make the established model more general, we assume that $N_p$ rotors are fixed on the robot, and we use $d_i$ (whose value is $\pm 1$) to denote the
rotating direction of the $i$th motor and $^{G}\boldsymbol{p}_{r,i}$ to denote its position in $\left\{\mathcal{G}\right\}$.}

If we use $\boldsymbol{f}_c=\left[f_1, \cdots, f_{N_p}\right]^T$ as input and try to express the thrust and torque in the rotor frame, the wrench becomes
\begin{subequations}
\label{eq:input_wrench}
\begin{align}
    {^{R_i}\boldsymbol{f}_i} &= \left[0, 0, f_i \right]^T, \quad f_i \in \left[f_{i,\min}, f_{i,\max}\right], \\[2pt]
    {^{R_i}\boldsymbol{\tau}_i} &= \left[0, 0, -d_i \ f_i \ \frac{k_q}{k_t} \right]^T.
\end{align}
\end{subequations}

Based on the coordinate systems, the tilt angle $\alpha_i$ influences the rotation from $\left\{\mathcal{R}_i\right\}$ to $\left\{\mathcal{E}_i\right\}$ along X axis, denoted as
\begin{equation}
    {^{E_i}_{R_i}\boldsymbol{R}} = \boldsymbol{R}_X\left(\alpha_i\right), \quad \alpha_i \in \left[\alpha_{i,\min}, \alpha_{i,\max}\right].
\end{equation}

In addition to the established torque, tilting a rotating propeller can generate an additional torque caused by gyroscopic effects. For aerial robots with small sizes, this effect is 2-3 orders of magnitude smaller \cite{bicego_nonlinear_2020} and can be treated as high-order terms to neglect. Then the resultant force and torque generated by propellers are derived as
\begin{subequations}
\label{eq:resultant_wrench}
\begin{align}
    {^G\boldsymbol{f}_u} = &\sum_{i=1}^{N_p}{^{G}_{E_i}\boldsymbol{R}} \ {^{E_i}_{R_i}\boldsymbol{R}} \ {^{R_i}\boldsymbol{f}_i}, \label{eq:fu} \\[2pt]
    \begin{split}
        {^G\boldsymbol{\tau}_u} = &\sum_{i=1}^{N_p}\left({^{G}_{E_i}\boldsymbol{R}} \ {^{E_i}_{R_i}\boldsymbol{R}} \ {^{R_i}\boldsymbol{\tau}_i} \right. \\
        &\left. + ^{G}\boldsymbol{p}_{r,i} \times {{^{G}_{E_i}\boldsymbol{R}} \ {^{E_i}_{R_i}\boldsymbol{R}} \ {^{R_i}\boldsymbol{f}_i}} \right), \label{eq:tau_u}
    \end{split}
\end{align}
\end{subequations}
where ${^{G}_{E_i}\boldsymbol{R}}$ can be obtained from geometric properties.

\subsubsection{Servo Model}

The servo can be modeled with angle, angular velocity, or even torque as input. Considering most low-cost servos only support angle control, we hereby model the servo's motion as a first-order model
\begin{equation}
    \label{eq:servo_model}
    \dot{\boldsymbol{\alpha}} = \frac{1}{t_{\rm servo}} \left( \boldsymbol{\alpha}_c - \boldsymbol{\alpha} \right),
\end{equation}
where $\boldsymbol{\alpha} = \left[\alpha_1, \alpha_2, \dots, \alpha_{N_p} \right]^T$ denotes the vector containing all servo angles, and $\boldsymbol{\alpha}_c = \left[\alpha_{c1}, \alpha_{c2}, \dots, \alpha_{cN_p} \right]^T$ is the symbol for servo angle commands.

\subsubsection{Rigid-Body Model}

Neglecting the aerodynamic drag during flying, the motion of the robot is caused only by gravitational force and rotor wrench. Using position ${^W\boldsymbol{p}}$, velocity $^W\boldsymbol{v}$, quaternion ${^W_G\boldsymbol{q}}$, and angular velocity ${^G \boldsymbol{\omega}}$ (angular velocity of $\left\{\mathcal{G}\right\}$ w.r.t $\left\{\mathcal{W}\right\}$ and expressed in $\left\{\mathcal{G}\right\}$) as states, a six-DoF rigid body dynamics can be established as follows
\begin{subequations}
\label{eq:rb}
\begin{align}
{^W\dot{\boldsymbol{p}}} &= {^W\boldsymbol{v}}, \label{eq:rb1} \\[5pt]
{^W\dot{\boldsymbol{v}}} &= \left({^{W}_{G}\boldsymbol{R}(\boldsymbol{q})} \ {^G\boldsymbol{f}_u} + {^W\boldsymbol{f}_d}\right)/ {m} + {^W\boldsymbol{{g}}}, \label{eq:rb2}\\[5pt]
{^W_G\dot{\boldsymbol{q}}} &= \frac{1}{2} \ {^W_G\boldsymbol{q}} \circ \mathcal{H}(^G\boldsymbol{\omega}), \label{eq:rb3}\\[5pt]
{^G\dot{\boldsymbol{\omega}}} &=\boldsymbol{{I}}^{-1}  \left(-{^G \boldsymbol{\omega}} \times\left({\boldsymbol{{I}}} \ {^G \boldsymbol{\omega}}\right)+{^G\boldsymbol{\tau}_u} + {^G\boldsymbol{\tau}_d}\right), \label{eq:rb4}
\end{align}
\end{subequations}
where $\boldsymbol{R}(\boldsymbol{q})$ denotes the rotation matrix converted from a quaternion, $\circ$ refers to quaternion multiplication, $\mathcal{H}(\cdot)$ means homogenizing a 3D vector $\mathcal{H}(\boldsymbol{p}) := [0, \boldsymbol{p}]^T$, ${^G\boldsymbol{f}_u}$ and ${^G\boldsymbol{\tau}_u}$ are expressed in (\ref{eq:fu}) and (\ref{eq:tau_u}), ${^W\boldsymbol{f}_d}$ and ${^G\boldsymbol{\tau}_d}$ are the force and torque caused by disturbances,
as well as $m$, ${^W\boldsymbol{{g}}}=[0,0,{-g}]^T$, and $\boldsymbol{I}\revised{=\texttt{diag} ({I}_{xx}, {I}_{yy}, {I}_{zz})}$ are the mass, the gravity vector, and the inertia matrix, respectively.


The whole model established above is utilized for control and simulation.
Note that this model is created for multirotors, and the tiltable-quadrotor is the special case when $N_p=4$.

\section{Control} \label{sec:control}


Unlike some cascade control algorithms, our proposed NMPC controller leverages the fully nonlinear dynamics as the control model, emphasizing a concise and elegant ``end-to-end" style. During this section, we first describe the generation of control reference in Sec. \ref{subsection:Planning}. Then this target is converted into a cost function inside a finite-time optimal control problem (OCP), which is the core concept of NMPC in Sec. \ref{subsection:NMPC}. One drawback of NMPC is its sensitivity to model error, which leads to a steady-state error during flight, \revised{especially influenced by the ground effect in the Z axis.} To solve this problem, a simple integral unit is introduced in Sec. \ref{subsection:I_term}. 

\subsection{Generation of Control Reference} \label{subsection:Planning}

Although NMPC can work with only $^W\boldsymbol{p}_r$ and $^W_G\boldsymbol{q}_r$, the optimizer can converge faster by computing reference states as comprehensively as possible. This calculation is worthwhile since the computation cost is trivial. Thus, all our subsequent experiments are conducted with the full-state reference, and the generation method is introduced here.

Compared with normal quadrotors, tiltable-quadrotors are able to independently track three-dimensional attitudes at the same time.
Hence, we need both reference position $^W\boldsymbol{p}_r$ and attitude $^W_G\boldsymbol{q}_r$ as control targets. Another difference in control reference comes from the number, where NMPC requires a sequence of reference points as the control target, not just one point for traditional feedback controllers.

The control task can be divided into two types: tracking a constant point or tracking a trajectory.
In our system, \revised{if the robot tracks a constant point at time $t$, then all reference points after $t$ are set the same}, and the reference velocity $^W\boldsymbol{v}_r$, acceleration $^W\dot{\boldsymbol{v}}_r$, angular velocity $^G\boldsymbol{\omega}_r$, and angular acceleration $^G\dot{\boldsymbol{\omega}_r}$ are set to zero. 
For tracking a trajectory, the reference points are calculated with a shifting prediction time interval, and $^W\boldsymbol{v}_r$, $^W\dot{\boldsymbol{v}}_r$, $^G\boldsymbol{\omega}_r$, and $^G\dot{\boldsymbol{\omega}_r}$ can be calculated by differentiating position and attitude.
Then the desired ${^G\boldsymbol{f}_{u,r}}$ and ${^G\boldsymbol{\tau}_{u,r}}$ can be calculated from (\ref{eq:rb2}) and (\ref{eq:rb4}), \revised{where the gyroscopic term is omitted for simplicity.}

On the basis of these states, we can obtain the reference thrust $f_{i,r}$ and servo angle $\alpha_{i,r}$ through control allocation. The relations between the reference wrench and a group of virtual input $\boldsymbol{z}$ can be expressed as
\begin{subequations}
\begin{align}
    &\left[^G\boldsymbol{f}_{u,r}, ^G\boldsymbol{\tau}_{u,r} \right]^T = \boldsymbol{A} \ \boldsymbol{z}, \quad \text{where} \\[5pt]
    \boldsymbol{z} = &\left[\begin{array}{c}
    f_{1,r,h} \\
    f_{1,r,v} \\
    \vdots \\
    f_{N_p,r,h} \\
    f_{N_p,r,v}
    \end{array}\right] = \left[\begin{array}{c}
    f_{1,r} \sin{\alpha_{1,r}} \\
    f_{1,r} \cos{\alpha_{1,r}} \\
    \vdots \\
    f_{N_p,r} \sin{\alpha_{N_p,r}} \\
    f_{N_p,r} \cos{\alpha_{N_p,r}}
    \end{array}\right],
\end{align}
\end{subequations}
where the allocation matrix $\boldsymbol{A}$ can be derived from (\ref{eq:fu}) and (\ref{eq:tau_u}) using symbolic computation tools. Then we can calculate $f_{i,r}$ and $\alpha_{i,r}$ as follows
\begin{subequations}
\begin{align}
\boldsymbol{z} &= \boldsymbol{A}^{\dagger} \ \left[^G\boldsymbol{f}_{u,r}, ^G\boldsymbol{\tau}_{u,r} \right]^T, \\[5pt]
f_{i,r} &= \sqrt{f_{i,r,h}^2 + f_{i,r,v}^2}, \\[5pt]
\alpha_{i,r} &= {\rm atan2}\left(f_{i,r,h}, f_{i,r,v}\right),
\end{align}
\end{subequations}
where $()^{\dagger}$ means the Moore–Penrose inverse operation. Note that $\boldsymbol{A}^{\dagger}$ is unchanged and only needs to be computed once.

\revised{As revealed by \cite{sun_comparative_2022}, NMPC has superiority in tracking dynamically infeasible trajectories. Hence, our requirement in this part is a middle-quality reference with a trivial computational burden, which has the potential for online replanning. The pose reference can be generated from parametric equations. If several pose points are given, the reference can be also obtained from the minimum-acceleration planning method \cite{mellinger_minimum_2011} or other evolutionary versions.}

\subsection{Nonlinear Model Predictive Control} \label{subsection:NMPC}

Nonlinear model predictive control converts the control problem into a constrained nonlinear optimization problem. For quadrotors, depending on the coordinate space, we can choose either a nonlinear model with linear constraints and a simple cost function, or a linear model with nonlinear constraints and a complex cost function. The former is easier to understand while the latter may have advantages in computational speed (please read \cite{greeff_flatness-based_2018} for discussions about normal quadrotors). Considering that the former one is more intuitive and has been deployed onboard successfully in many recent research \cite{sun_comparative_2022, li_nonlinear_2023}, we select the nonlinear model version.

We select the state as $\boldsymbol{x} = \left[{^W\boldsymbol{p}}, {^W\boldsymbol{v}}, {^W_G\boldsymbol{q}}, {^G\boldsymbol{\omega}}, {\boldsymbol{\alpha}}\right]^T$ and the control input as $\boldsymbol{u} = \left[\boldsymbol{f}_c, \boldsymbol{\alpha}_c\right]^T$. Given the reference $\boldsymbol{x}_r$, $\boldsymbol{u}_r$, we define the state error as $\overline{\boldsymbol{x}} = \left[ \overline{\boldsymbol{p}}, \overline{\boldsymbol{v}}, \mathcal{V}(\boldsymbol{q}_e), \overline{\boldsymbol{\omega}}, \overline{\boldsymbol{\alpha}} \right]^T$ and the control input error as $\overline{\boldsymbol{u}} = \left[\overline{\boldsymbol{f}_c}, \boldsymbol{\alpha}_c - \boldsymbol{\alpha} \right]^T$, where the over-line symbol denotes $\overline{(\cdot)}=(\cdot) - (\cdot)_r$ if no special explanation, $\mathcal{V}(\cdot)$ represents the vector part of a quaternion $\mathcal{V}(\boldsymbol{q}) := \left[q_x, q_y, q_z\right]^T$, and ${\boldsymbol{q}}_e=\boldsymbol{q} \circ \boldsymbol{q}_r^{-1}$ represents the quaternion error.
Note that the error term for the servo angle command is defined as $\overline{\boldsymbol{\alpha}}_c=\boldsymbol{\alpha}_c - \boldsymbol{\alpha}$ instead of the error w.r.t. reference $\overline{\boldsymbol{\alpha}}_c'=\boldsymbol{\alpha}_c - \boldsymbol{\alpha}_r$, aiming to penalize the right-hand of (\ref{eq:servo_model}) to avoid the sudden change of servo commands.

Then the optimal control problem in NMPC can be formulated as a \revised{nonlinear least-square problem}
\begin{subequations}
\begin{align}
    \underset{\boldsymbol{x}_k,\boldsymbol{u}_k}{\rm minimize \ } \quad &\sum\limits_{k=0}^{N-1}\left(\overline{\boldsymbol{x}}^T_k \boldsymbol{{Q}} \overline{\boldsymbol{x}}_k + \overline{\boldsymbol{u}}^T_k \boldsymbol{{R}} \overline{\boldsymbol{u}}_k\right) + \overline{\boldsymbol{x}}^T_N \boldsymbol{{Q}}_N\overline{\boldsymbol{x}}_N, \label{eq:cost} \\
    {\rm subject \  to} \quad 
    &\boldsymbol{x}_{k+1} = \boldsymbol{f}\left(\boldsymbol{x}_{k}, \boldsymbol{u}_{k}\right), \quad k=0:N-1, \label{eq:dyn_constraint} \\
    &\boldsymbol{x}_0 =\hat{\boldsymbol{x}}, \label{eq:init_constraint} \\
    &\left\|v_{x,y,z}\right\| \leq v_{\rm limit}, \quad \left\|\omega_{x,y,z}\right\| \leq \omega_{\rm limit}, \label{eq:state_constraint} \\
    &\boldsymbol{u}_{\min} \leq \boldsymbol{u}_k \leq \boldsymbol{u}_{\max}, \label{eq:input_constraint}
\end{align}
\end{subequations}
where $\boldsymbol{Q}$, $\boldsymbol{R}$, $\boldsymbol{Q}_N$ are positive diagonal matrices representing weights for state cost, control energy cost, and terminal cost, respectively. 
The (\ref{eq:dyn_constraint}) indicates the dynamics constraint, where $\boldsymbol{f}(\cdot)$ refers to the full tiltable-quadrotor model established from (\ref{eq:input_wrench}) to (\ref{eq:rb}). This model is discretized by the \textit{fourth-order Runge–Kutta} method with $t_{\rm integ}$ integrating time.
The (\ref{eq:init_constraint}) denotes the initial value constraint, a critical component for feedback in NMPC, wherein $\hat{\boldsymbol{x}}$ represents the estimated state from the estimator. 
Finally, the (\ref{eq:state_constraint}) and (\ref{eq:input_constraint}) refer to the state and input constraints, respectively, which are mainly decided from safety concerns and physical limits.

\revised{After calculation, the first element of the optimized sequence $\boldsymbol{u}^*$ is transmitted to a low-level autopilot for execution:}
\begin{equation}
    \revised{\boldsymbol{u}_{\rm now} = \boldsymbol{u}^*_0.}
\end{equation}

In implementation, we leverage several typical techniques in the NMPC community to accelerate computation. Specifically, \textit{warm-starting}, \textit{real-time iteration (RTI)}, and \textit{multiple-shooting} are adopted. The \textit{warm-starting} tries to accelerate by giving an initial guess near the final solution, where the guess comes from the last round's result. The RTI only calculates the sequential quadratic programming (SQP) once in one control iteration, preferring speed instead of optimality. Finally, the \textit{multiple-shooting} divides the original OCP into several smaller optimization problems and tries to solve them in parallel. We recommend \cite{gros_linear_2020} to interested readers for more details.

\subsection{\revised{Integral Term for Ground Effect}} \label{subsection:I_term}

\revised{When rotor-based drones fly near the ground, the aerodynamic interaction between the drone's propellers and the environment can increase the lift, resulting in higher force in the Z axis. We empirically find that the model error caused by this phenomenon is much larger than other axes, so we adopt an integral term as follows to compensate for the Z error}
\begin{equation}
        \revised{f_{d,z} = \text{ITerm}\left({^W\hat{p}_z} - {^Wp_{z,r}} \right).}
\end{equation}
The digital version of the integral term $u[k+1] = \text{ITerm}(e[k+1])$ with trapezoidal rule and anti-windup is given as \cite{beard_small_2012}
\begin{align}
    \begin{split}
        I'[k+1] &= I[k] + \frac{t_s}{2} \left(e[k] + e[k+1] \right), \\
        u'[k+1] &= k_I \ I'[k+1], \\
        u[k+1] &= \max\left( \min\left(u'[k+1], u_{\max}\right), u_{\min}\right), \\
        I[k+1] &= I'[k+1] + \frac{1}{k_I}  \left(u[k+1] - u'[k+1] \right).
    \end{split}
\end{align}

The compensating \revised{force ${^W\boldsymbol{f}_d}=\left[0, 0, f_{d,z}\right]^T$} is calculated before each NMPC round and then transmitted as parameters into the optimization process through (\ref{eq:rb2}).


\setlength{\dbltextfloatsep}{8pt plus 1.0pt minus 2.0pt}
\begin{figure}[t] 
    \centering
    \subfloat[\revised{Pose tracking for NMPC without servo and thrust model}]{
        \includegraphics[trim=0cm 0cm 0 0.1cm,clip,width=3.5in]{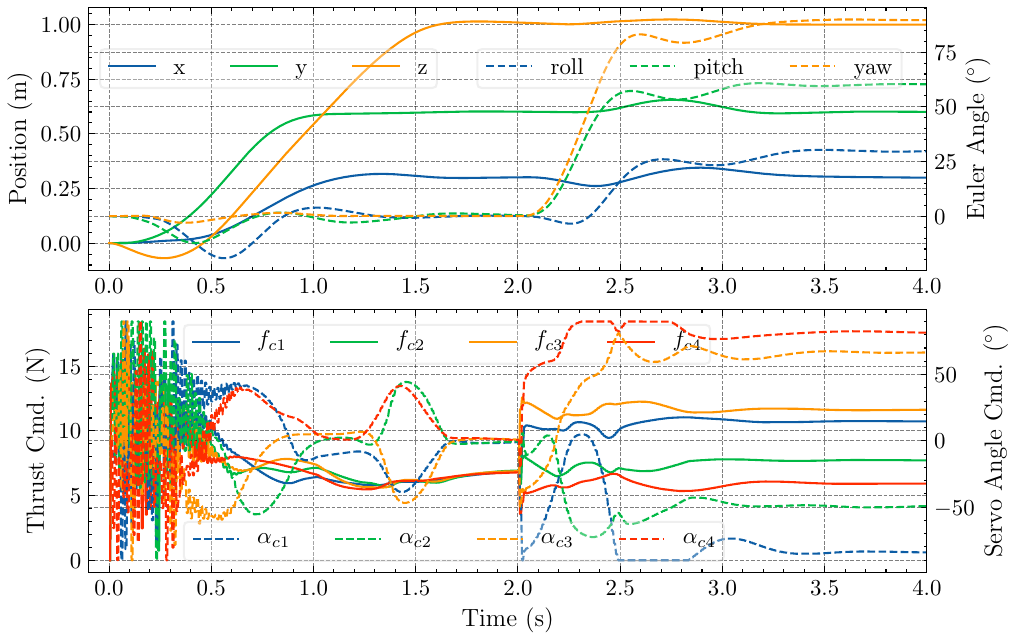}
        \label{fig:pysim_wo_servo_wo_rotor}
    }\\ \vspace{-1mm}
    \subfloat[\revised{Pose tracking for NMPC with servo and without thrust model.}]{
        \includegraphics[trim=0cm 0cm 0 0.1cm,clip,width=3.5in]{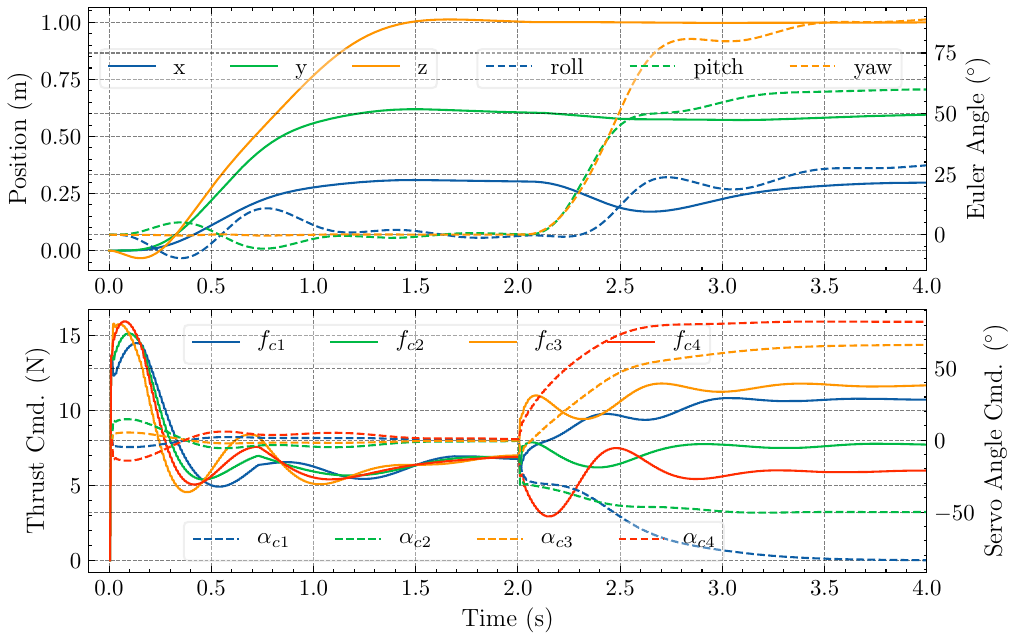}
        \label{fig:pysim_w_servo_wo_rotor}
    }\\ \vspace{-1mm}
    \subfloat[\revised{Pose tracking for NMPC with servo and thrust model}]{
        \includegraphics[trim=0cm 0cm 0 0.1cm,clip,width=3.5in]{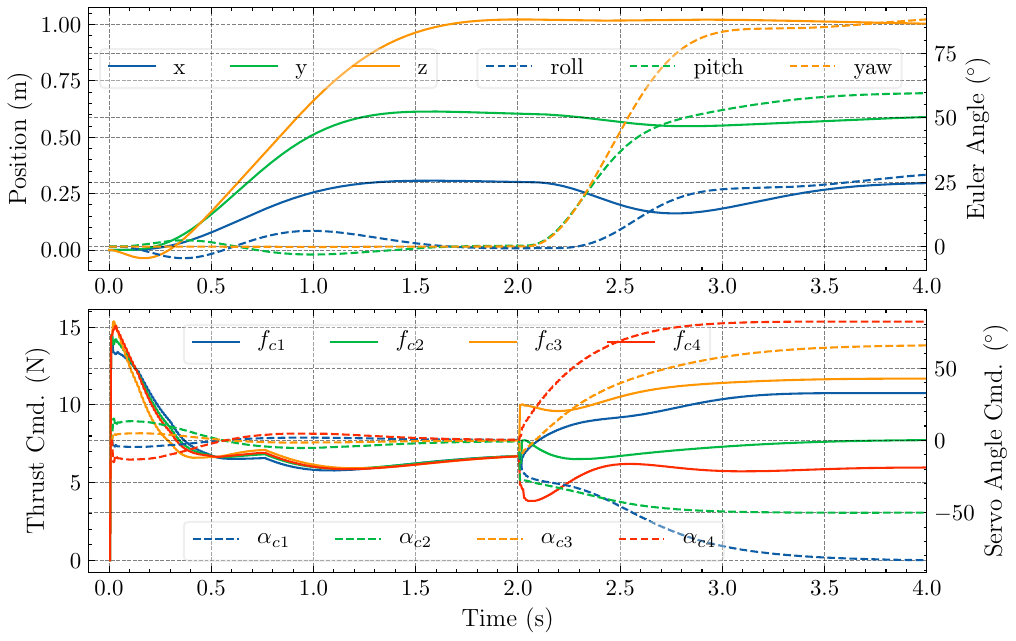}
        \label{fig:pysim_w_servo_w_rotor}
    }
    \caption{\revised{Comparative analysis of NMPC considering different models in an ideal simulation, where the control targets are described in the main text. The drastic command oscillation in (a) leads to optimization errors in noisy simulations such as Gazebo, and incorporating a servo model in (b) significantly mitigates this phenomenon. From (b) and (c), adding a thrust model merely results in a smoother tracking performance. The result of only considering the thrust is similar to (a).}}
    \vspace{-0.1cm}
    \label{fig:pysim_sim}
\end{figure}

\section{Simulation} \label{sec:simulation}

\revised{In this section, we perform simulations to reveal the impact of the servo model on the NMPC performance.
To eliminate the influence of model error, disturbance, and other unknown effects, we implement an ideal simulation environment without model error and noise.}

\revised{When simulating, the target is first to track a position of $\boldsymbol{p}_r=[0.3, 0.6, 1.0]^T$ from $t=0$ s, then to track an attitude of $\boldsymbol{q}_r={\rm RPY2Quat}\left(30^\circ, 60^\circ, 90^\circ\right)^T$ from $t=2$ s\revised{, where ${\rm RPY2Quat}(\cdot)$ denotes converting Euler angles to quaternions}. These targets are given as step signals to represent the challenging dynamically infeasible cases.
In the subsequent comparison, we choose the control frequency as 100 Hz and the simulation frequency as 200 Hz to align with the real-world experiments. To make the model more realistic, besides the servo-integrated nonlinear model (\ref{eq:input_wrench})-(\ref{eq:rb}), we also consider the thrust model as a first-order system similar to (\ref{eq:servo_model}), where the time constant $t_{\rm thrust}$ is set from the identification result (Fig. \ref{fig:rotor_model_sysid}). The disturbances ${^W\boldsymbol{f}_d}$ and ${^G\boldsymbol{\tau}_d}$ are all set to zero, and other parameters are set the same as in Table \ref{tab:control_params}. The simulation results are depicted in Fig. \ref{fig:pysim_sim}.}



\subsection{\revised{The Influence of Servo Model on Optimization Convergence}}

\revised{It is natural to think that adding a model increases the OCP complexity, resulting in higher difficulty in convergence.
However, as revealed by Fig. \ref{fig:pysim_wo_servo_wo_rotor} and Fig. \ref{fig:pysim_w_servo_wo_rotor}, the servo-integrated NMPC achieves less oscillation and faster optimization convergence.
We believe this is due to the narrower search space introduced implicitly by the first-order servo model. Without this model, the next servo angle command can be any value within the physical limits. With this model, the command is limited in the nearby area of the real servo angle, and this extra limitation accelerates the convergence. 
Although the no-servo version achieves optimization convergence in the end, we report it diverging and crashing in noise-existing simulation, not mentioned in the real world. Hence, the inclusion of the servo model is a must for real-world usage.}

\revised{We also observe that increasing the servo angle's weight to a large number (above 50) can eliminate the oscillation. However, this way relies heavily on the accuracy of $\boldsymbol{\alpha}_r$ and leads to unacceptable error when tracking infeasible trajectories.}

\subsection{\revised{The Influence of Servo Model on Flight Performance}}

\revised{On the basis of optimization convergence, considering the servo dynamics can reduce the fluctuations and hence increase the flight performance as revealed by  Fig. \ref{fig:pysim_wo_servo_wo_rotor} and Fig. \ref{fig:pysim_w_servo_wo_rotor}.}

\revised{It is worth considering whether the thrust model should be incorporated when its time constant is comparable to that of the servo. Here we emphasize that, even if their time constants are similar, the servo delay still has a more significant influence on the flight performance than thrust, and the reasons come from their different working ranges and servo angle's nonlinearity. During the mild flight, since the rotors of tiltable-quadrotors are always tilted upward, the relative working range of thrust is much smaller than the servo angle. This fact can be observed from Fig. \ref{fig:pysim_w_servo_wo_rotor}: after 2 s, the servo angle $\alpha_{c1}$ decreases from $0^\circ$ to $-80^\circ$ while the thrust $f_{c1}$ increases from $7$ N to $11$ N. If they have the same time constant and all left 20\% to reach the target, the lagging angle and thrust are $(-80^\circ - 0^\circ) \times 20\% = -16^\circ$ and $(11 - 7)) \times 20\% = 0.8$ N, respectively. Mathematically, the wrench (\ref{eq:fu}) and (\ref{eq:tau_u}) are the sum of terms such as $c_x \sin(\alpha_i) f_i$ or $c_y \cos(\alpha_i) f_i$, where $c$ denotes the constant term. Thus, a $-16^\circ$ lag causes $(\cos(-64^\circ)-\cos(-80^\circ))/\cos(-80^\circ) \times 100\%=152.45\%$ relative error, while the $0.8$ N only causes $(10.2-11)/11 \times 100\%=-7.27\%$ relative error. For other ranges where the cosine term is trivial, the sine term is important. The greater importance of servo angle than thrust is also verified in Fig. \ref{fig:pysim_sim}, where they have unequal influences on performance.}

\revised{Although considering the thrust model can theoretically improve performance, focusing solely on the servo model is easier for implementation. Adding the thrust model to NMPC needs more computation, and also requires extra state-feedback of thrust, which can be calculated from rotor speed through (\ref{eq:rotor_dynamics}). Many servos support angle measurement, but few ESCs can measure rotor speed at high frequencies. Therefore, we recommend only considering the servo model at the beginning. If oscillations are observed as in Fig. \ref{fig:pysim_w_servo_wo_rotor}, modeling the thrust should then be considered.}

\section{Experiments} \label{sec:experiments}

\subsection{Robot Platform}

\setlength{\textfloatsep}{8pt plus 1.0pt minus 2.0pt}
\begin{figure}[b]
    \centerline{\includegraphics[trim=0 0.1cm 0 0,clip,width=3.5in]{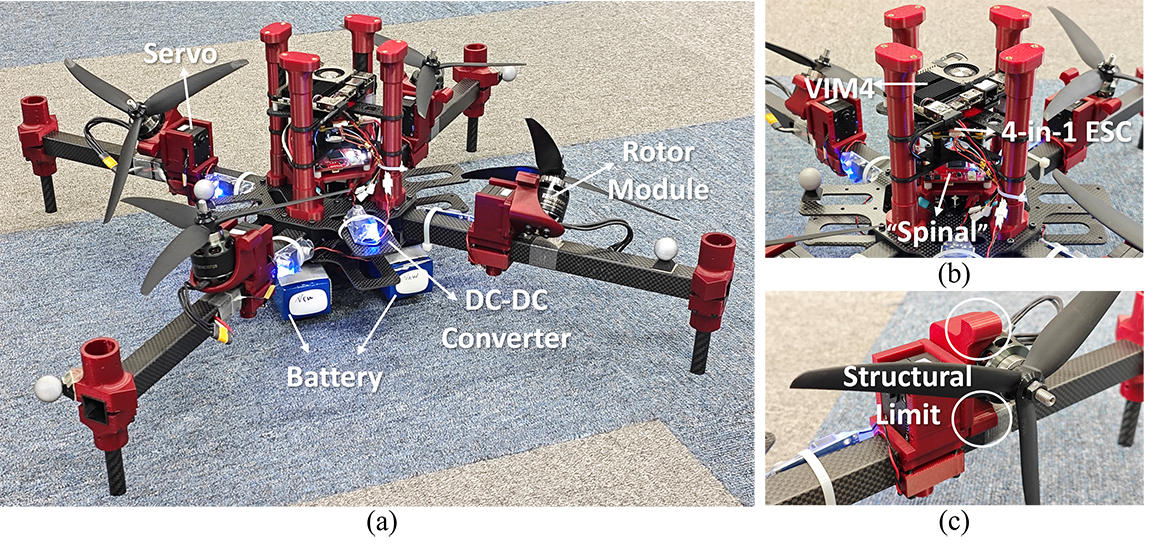}}
    \vspace*{-3mm}
    \caption{(a) Snapshot of our self-build tiltable-quadrotor. (b) An onboard computer VIM4, a 4-in-1 electronic speed controller (ESC), and a self-designed flight control unit ``Spinal" are centrally mounted on the robot's body. (c) The feasible servo angle is structurally limited to within $\pm \pi/2$.}
    \label{fig:robot_detail}
\end{figure}

\begin{figure}[t]
    \centerline{\includegraphics[trim=6.5cm 6cm 8.5cm 4.5cm,clip,width=3.4in]{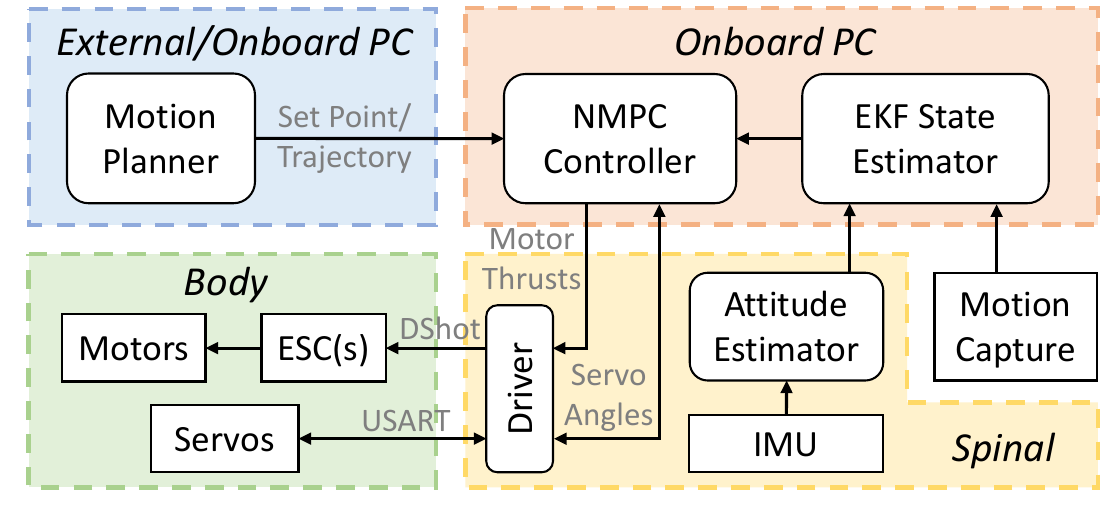}}
    \vspace*{-3mm}
    \caption{Diagram of information flow among main modules within our system, where normal rectangles denote hardware modules and rounded rectangles indicate software modules.}
    \label{fig:workflow_detail}
\end{figure}

We made a tiltable-quadrotor as shown in Fig. \ref{fig:robot_detail} to verify the proposed approach, \revised{and the basic parameters are listed in Table \ref{tab:control_params}.}
The main modules and their communications are illustrated in Fig. \ref{fig:workflow_detail}, where the Robot Operating System (ROS) is used for communication. The NMPC controller is running in a Khadas VIM4 onboard computer, which has a 2.2 GHz Quad-Core ARM Cortex-A73 and a 2.0 GHz Quad-Core Cortex-A53 CPU. Then it sends control commands to the self-designed flight control unit ``Spinal", which employs an STM32H7 series processor and has been employed in many previous works \cite{zhao_design_2023,nishio_design_2024}. Next, ``Spinal" sends signals to control actuators. Specifically, we choose a T-Motor F55A PROII 6S 4IN1 ESC with DShot protocol to power four T-Motor AT2814 KV900 motors. Each motor is equipped with a 3-blade 9045 propeller. We select Kondo KRS-3302 as servos.

\setlength{\textfloatsep}{8pt plus 1.0pt minus 2.0pt}

For state estimation, an IMU is installed on ``Spinal", and its data are filtered in an attitude estimator to obtain roll and pitch angles as well as attitude velocity. In addition, we use an OptiTrack Motion Capture system to get position and attitude. These measurements are input to an EKF-based state estimator to calculate $^W\hat{\boldsymbol{p}}$, $^W\hat{\boldsymbol{v}}$, $^W_G\hat{\boldsymbol{q}}$, and $^G\hat{\boldsymbol{\omega}}$. The servo angle $\hat{\boldsymbol{\alpha}}$ is directly obtained by the Kondo servo without filtering. These estimated states are sent to NMPC for control.

\subsection{\revised{Parameter Identification}}

\revised{Our parameter identification experiments can be separated into three categories: geometry and inertial parameters, rotor parameters, as well as servo parameters.}

The geometric parameters for the $i$th rotor $^G\boldsymbol{p}_{r,i}$ were directly obtained from the CAD model. The mass parameter was weighted using a weighing scale, and the inertial matrix was identified by the bifilar-pendulum method \cite{jardin_optimized_2009}. 

The rotor parameters $k_t$, $k_q$, $f_{i,\min}$, $f_{i,\max}$ as well as the mapping between command and thrust were identified by an ATI six-axis force/torque sensor fixed on a testbed. \revised{To make the result closer to reality, we 3D-printed a servo structure and mounted it near the rotor with $0^\circ$ tilting. Compared to directly fixing the rotor on the force sensor, we observed a 19.53\% thrust loss for the same command.}

\setlength{\dbltextfloatsep}{8pt plus 1.0pt minus 2.0pt}
\begin{figure}[b] 
    \centering
    \vspace{-5mm}
    \subfloat[Result for servo identification]{\hspace{-0.2cm}
        \includegraphics[trim=0.1cm 0 0 0,clip,width=1.7in]{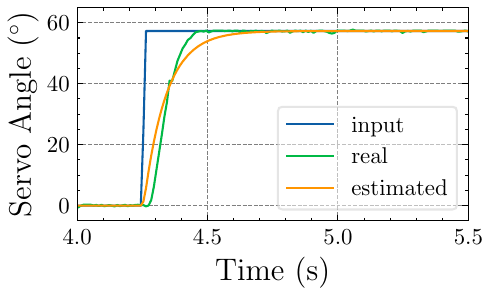}
        \label{fig:servo_model_sysid}
    }
    \subfloat[\revised{Result for rotor identification}]{\hspace{-0.2cm}
        \includegraphics[trim=0.1cm 0 0 0,clip,width=1.7in]{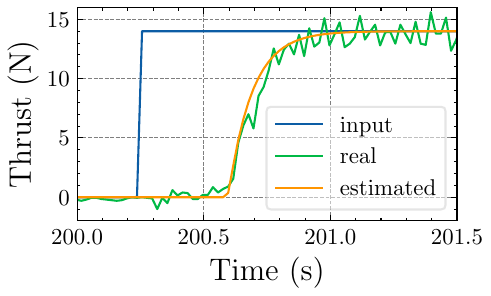}
        \label{fig:rotor_model_sysid}
    }\vspace{-1mm}
    \caption{For servos, the identified time constant is 0.0859 s, achieving an average accuracy of 93.4\%. \revised{For rotors, the time constant is 0.0942 s with a 0.35-s dead time, where this large dead time is only observed when starting from zero speed. Thus, the time constants for servo and thrust can be similar.}}
    \label{fig:sysid_first_order}
\end{figure}

To identify the servo model, we removed the propellers for safety, commanded the motor to rotate at 60\% throttle, and sent a step signal to stimulate the servo to 1 rad ($57.30^\circ$). The data were collected using \texttt{rosbag}, and then the time constant of the servo model was identified by the MATLAB System Identification Toolbox, where Prediction Error Minimization is used as the evaluation method. The final result is the average of four servos, with one servo's result as shown in Fig. \ref{fig:sysid_first_order}.
Although this no propeller setting may introduce some errors, this accuracy is enough for real-world flight.



\subsection{Tracking Test}

The tool acados \cite{verschueren_acadosmodular_2022} is utilized to solve the NMPC problem, where we use \textit{Partial Condensing HPIPM} as QP solver and \textit{Explicit Runge-Kutta} as integrator. The control parameters are listed in Table \ref{tab:control_params}. We conducted three experiments to evaluate the proposed algorithm. Initially, the tiltable-quadrotor took off and then was poked by a stick to test disturbance rejection. Next, we carried out an experiment to track several pose points, aiming to evaluate the performance of physically infeasible references. Finally, the robot was controlled to track \revised{a 6-DoF pose trajectory with different speeds.}

\begin{table}[t]
\setlength{\abovecaptionskip}{0pt} 
\setlength{\belowcaptionskip}{-1pt}
\centering
\caption{Model \& Control Parameters}
\begin{tabular*}{\linewidth}{@{\extracolsep{\fill}} c c|c c|c c }
 \toprule
 \textbf{Param.} & \textbf{Value} & \textbf{Param.} & \textbf{Value} & \textbf{Param.} & \textbf{Value} \\
 \midrule
  \revised{Wheelbase} & \revised{$0.4$ m} & \revised{$m$} & \revised{$2.773$ kg} &  \revised{$\alpha_{\rm limit}$} & \revised{$\pm \pi/2$}  \\
 \revised{$I_{xx}$} & \revised{$0.0417$} & \revised{$I_{yy}$} & \revised{$0.0395$} & \revised{$I_{zz}$} & \revised{$0.0707 \ \text{kg m}^2$} \\
$N_p$ & $4$ & $k_q/k_t$ & $0.0153$ & $t_{\rm servo}$ & $0.0859$ s  \\
 \midrule
 $N$ & $20$ & $t_{\rm integ}$ & $0.1$ s & $t_s$ & $0.01$ s \\
 $Q_{p,\rm xy}$ & $300$ & $Q_{p,\rm z}$ & $400$ & $Q_{\rm v,xy}$ & $10$ \\
 $Q_{v,\rm z}$ & $10$ & $Q_{q,\rm xy}$ & $300$ & $Q_{\rm q,z}$ & $600$ \\
 $Q_{\omega,\rm xy}$ & $5$ & $Q_{\omega,\rm z}$ & $5$  & $Q_{\alpha}$ & $\revised{2}$  \\
 $R_{f}$ & $\revised{2}$ & $R_{\alpha}$ & $250$ & & \\
 $v_{\rm limit}$ & $\pm1$ m/s & $\omega_{\rm limit}$ & $\pm6$ rad/s & $\alpha_{i,{\rm limit}}$ & $\pm \pi/2$ \\
 $f_{i,\min}$ & $0$ N & $f_{i,\max}$ & $30$ N & $\alpha_{ci,{\rm limit}}$ & $\pm \pi/2$ \\
 \midrule
  $k_{I,z}$ & $5$ & $f_{d,\rm limit}$ & $5$ N &  & \\
 \bottomrule
\end{tabular*}
\label{tab:control_params}
\vspace{-1mm}
\end{table}

\subsubsection{\revised{Takeoff, Anti-Disturbance, and Landing}}

\begin{figure}[t]
    \rightline{\includegraphics[trim=0 0 0 0,clip,width=3.4in]{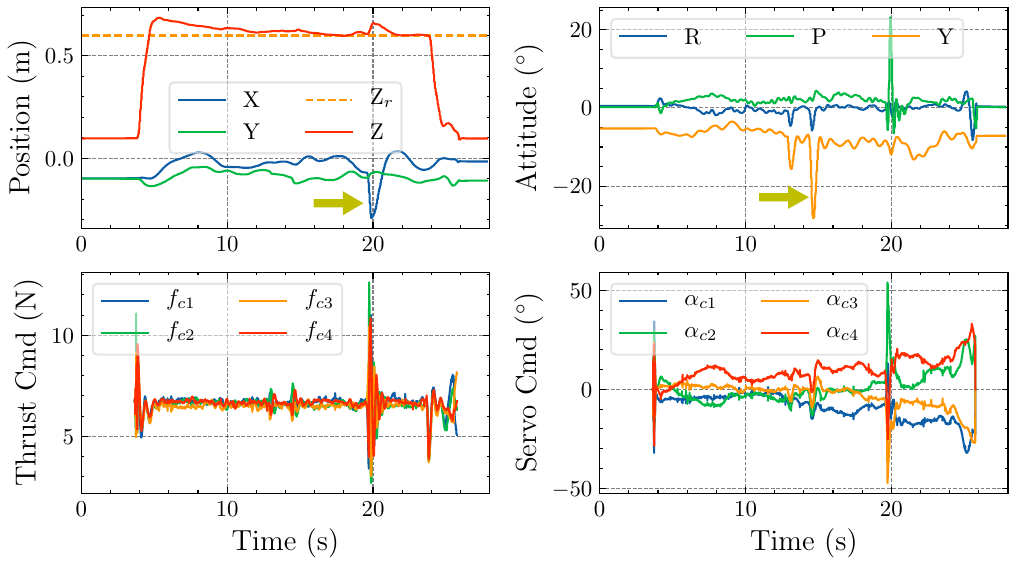}}
    \vspace*{-3mm}
    \caption{\revised{Pose data from the takeoff, anti-disturbance, and landing experiment. A disturbance is given to yaw at about 14.5s (the yellow arrow in the left-upper figure), and then another disturbance is given to X at about 20s (the yellow arrow in the right-upper figure).}}
    \label{fig:takeoff_data}
\end{figure}

Takeoff is the basis for all other experiments. \revised{During takeoff, the position command with the current X-Y position and $p_{r,z}=0.6$ m was sent to the robot.} After takeoff, the robot entered a hovering state and was then disturbed by a stick. \revised{After that, the robot landed.} The data are shown in Fig. \ref{fig:takeoff_data}.

\revised{From Fig. \ref{fig:takeoff_data}, the robot can achieve the target height with a 14.83\% overshoot, and then the integral term slowly corrects the Z position to the reference within 6 s. After giving disturbances to the robot, it can recover to the hovering state within 1 s, demonstrating the robustness of the proposed controller. After the disturbances ($t>21$ s), the tilting angles of about $10^\circ$  still exist, possibly implying that the optimizer has fallen into a local minimum. More analysis can be done in the future.}

\subsubsection{Set-Pose Tracking}

\begin{figure}[bt]
    \centerline{\includegraphics[trim=0 0 0 0,clip,width=3.5in]{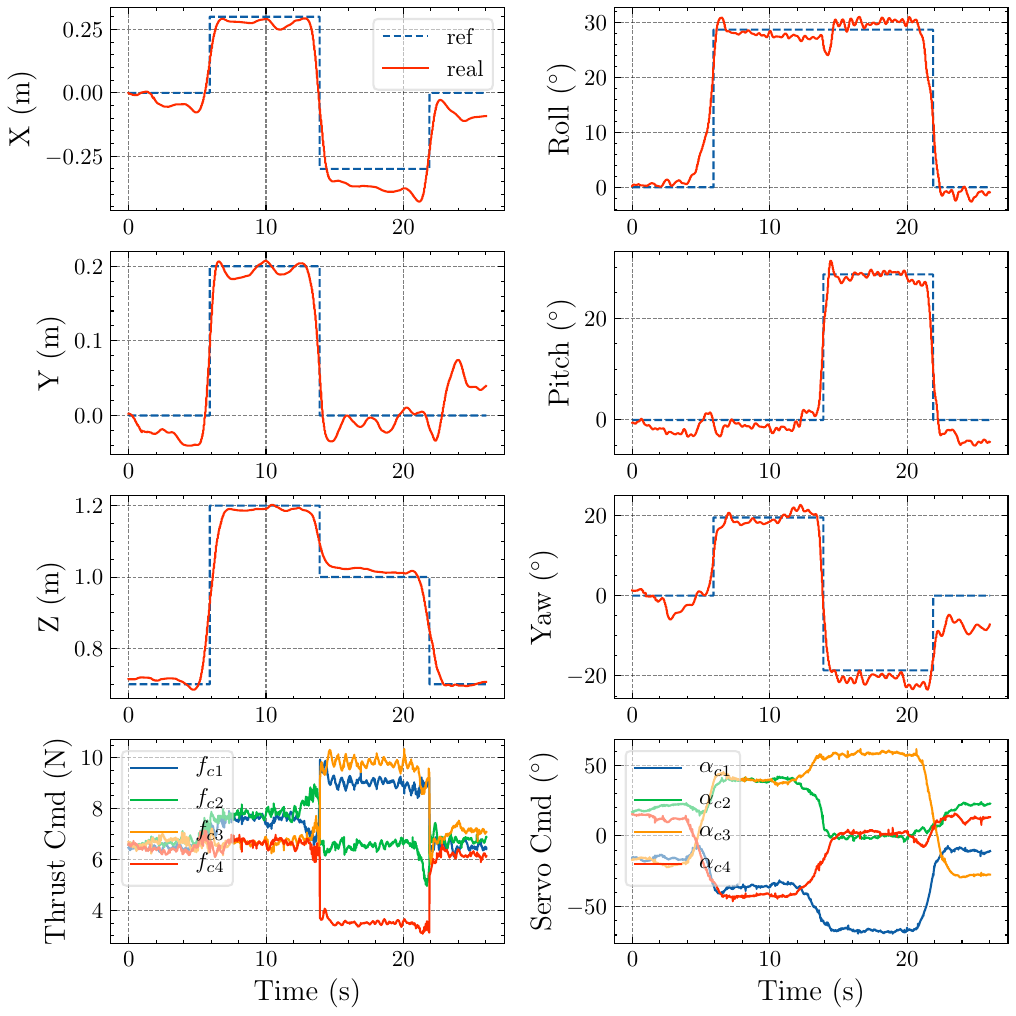}}
    \vspace*{-3mm}
    \caption{\revised{Pose data from the set point tracking experiment. The robot tracks two pose points and finally goes back to the starting point.}}
    \label{fig:set_point}
\end{figure}

The capability of tracking static points was tested, and the reference in each DoF can be seen as a step signal. Initially, the pose point $\boldsymbol{p}_r[\text{m}]=\left[0.3,0.2,1.2\right]^T, \ \boldsymbol{q}_r={\rm RPY2Quat}\left(0.5,0.0,0.3\right)^T$ was sent to the robot. After eight seconds, another pose point $\boldsymbol{p}_r[\text{m}]=\left[-0.3,0.0,1.0\right]^T, \ \boldsymbol{q}_r={\rm RPY2Quat}\left(0.5,0.5,-0.3\right)^T$ was sent. After another eight seconds, the robot was commanded to the starting point. The result is plotted in Fig. \ref{fig:set_point}. \revised{Note that we use radian in code but degree in figures for readability.}

From Fig. \ref{fig:set_point}, the robot can track the position and attitude changes simultaneously due to the non-simplification of the nonlinear robot model. \revised{The tracking RMSE for all directions are X: 0.072 m, Y: 0.029 m, Z: 0.044 m; Roll: 3.250$^\circ$, Pitch: 2.810$^\circ$, Yaw: 4.342$^\circ$. Some oscillations in Y exist from 16 to 22 s, possibly due to aerodynamic disturbances from environments. In addition, the X position has an error of about 0.08 m from 14 to 22 s, indicating the existence of model error. Overall, we demonstrate the capability of setpoint tracking.}






\setlength{\textfloatsep}{8pt plus 1.0pt minus 2.0pt}

\setlength{\textfloatsep}{8pt plus 1.0pt minus 2.0pt}

\subsubsection{Pose Trajectory Tracking}

\begin{figure}[tb]
    \centerline{\includegraphics[trim=0 0 0 0,clip,width=3.5in]{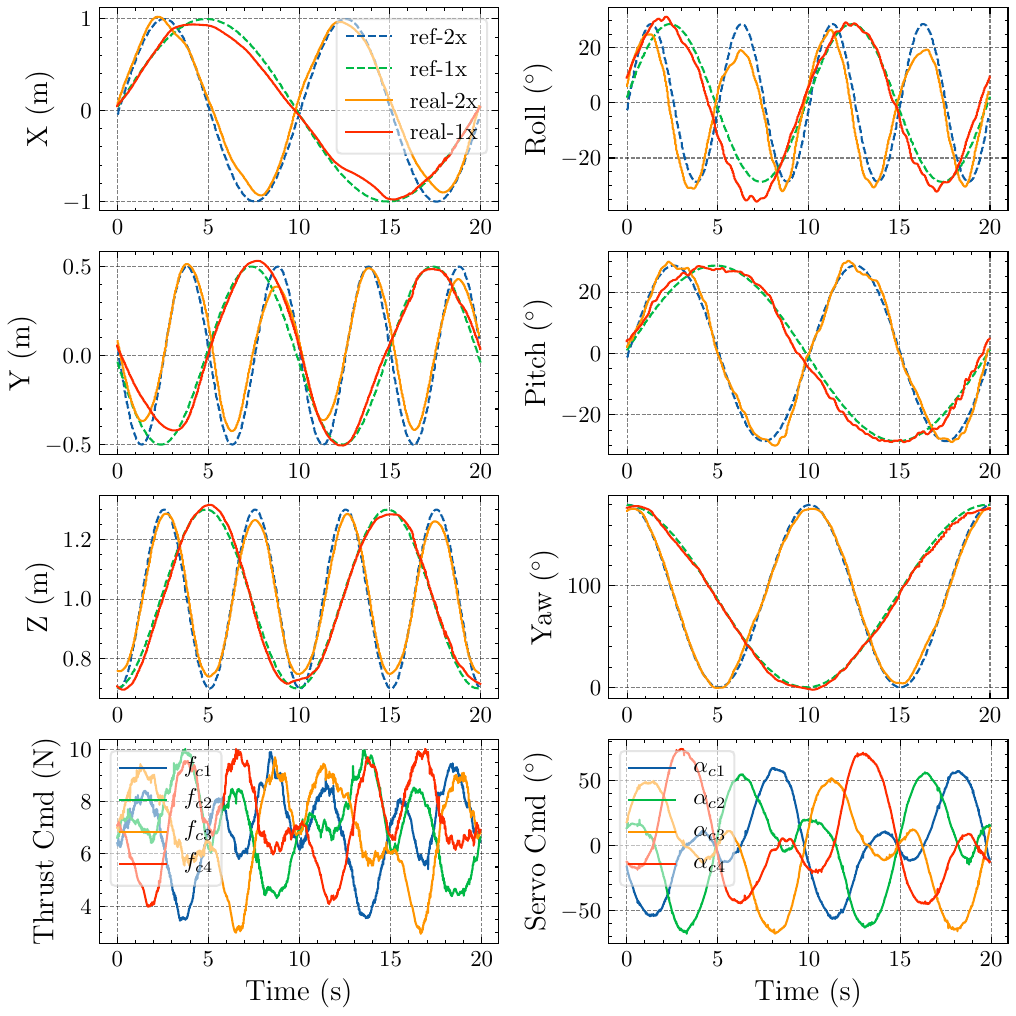}}
    \vspace*{-3mm}
    \caption{\revised{Pose data for tracking one trajectory with different speeds, where 1x means using 20 s and 2x means using 10 s to finish. The plotted commands are for 2x trajectory. The actual servo angles follow their commands with a slight delay.}}
    \label{fig:traj}
\end{figure}

Finally, a pose trajectory was sent to the robot. Let $\omega=2\pi/T$, then the position $\boldsymbol{p}_r$[m] was set as $p_x(t) = \cos(\omega t),\  p_y(t) = \sin(2 \omega t)/2,\  p_z(t) = 0.3 \sin(2 \omega t + \pi/2) + 1.0$, as well as the attitude $\boldsymbol{q}_r$ was set from Euler angles as $roll(t) = - \sin(2 \omega t)/2,\  pitch(t)=0.5\cos(\omega t),\ yaw(t)=\pi/2 \cdot \sin(\omega t + \pi) + \pi/2$. \revised{We set $T=20$ s (1x) and $T=10$ s (2x) for one trajectory with different difficulties}, where the tracking results are displayed in Fig. \ref{fig:traj}, and the real-world flight snapshot is presented in Fig. \ref{fig:flight}.

\setlength{\textfloatsep}{8pt plus 1.0pt minus 2.0pt}

\revised{The tracking RMSE for 1x trajectory are X: 0.071 m, Y: 0.067 m, Z: 0.018 m; Roll: 4.934$^\circ$, Pitch: 2.023$^\circ$, Yaw: 2.789$^\circ$. RMSE for 2x trajectory are X: 0.103 m, Y: 0.085 m, Z: 0.029 m; Roll: 6.740$^\circ$, Pitch: 1.857$^\circ$, Yaw: 3.622$^\circ$, worse than 1x case.
The tracking error in some quick turns (e.g., at 6.5 s for Roll-2x) may be due to model error or physical infeasibility, which can be resolved in the future. Overall, we demonstrate the feasibility of the proposed controller for trajectory tracking.}



\section{Conclusion} \label{sec:conclusion}

In this article, we proposed an NMPC-based control framework for tiltable-quadrotors. Leveraging a fully nonlinear model with servo dynamics, the method directly generated rotor thrust and servo angle as control inputs. We found in the simulation that the inclusion of servo dynamics not only enhanced the control performance but also assisted in the \revised{optimization} convergence.
Finally, the algorithm was verified in the real world using a self-made robot.

\revised{In the future, we plan to implement a disturbance observer within the system to achieve offset-free tracking. Additionally, the proposed controller can be generalized to other aerial robots with tilting structures. We also prepare to extend our approach to handle physical interactions.}







\ifCLASSOPTIONcaptionsoff
  \newpage
\fi


\bibliographystyle{bibtex/IEEEtran}
\bibliography{bibtex/IEEEabrv, bibtex/my_config, bibtex/references }


\end{document}